\documentclass[journal]{IEEEtran}
\usepackage{algorithm}
\usepackage[english]{babel}
\usepackage{blindtext}
\usepackage{algorithmic}
\usepackage{color, soul}

\usepackage{multirow}
\usepackage{xcolor}
\usepackage{amsfonts}
\usepackage{overpic}
\usepackage{epsfig}
\usepackage{graphicx}
\usepackage{amsmath}
\usepackage{amsmath,bm}
\usepackage{amssymb,subfigure,cases}
\usepackage{color,hyperref}
\usepackage{caption}
\usepackage{epstopdf}
\usepackage{amsmath}
\usepackage{makecell}
\usepackage{flushend}

\ifCLASSINFOpdf
\else
\fi
\definecolor{darkblue}{rgb}{0.0,0.0,1.0}
\hypersetup{colorlinks,breaklinks,
            linkcolor=darkblue,urlcolor=darkblue,
            anchorcolor=darkblue,citecolor=darkblue}

\begin{document}
%\linenumbers
%\switchlinenumbers

\title{Pairwise Comparison Network for Remote Sensing Scene Classification}
\vspace{+0.2cm}
\author{Yue~Zhang,
        Xiangtao Zheng,~\IEEEmembership{Member,~IEEE},
        and Xiaoqiang Lu,~\IEEEmembership{Senior Member,~IEEE}
       % \vspace{-0.93cm}
\vspace{+0.3cm}
\thanks{ This work was supported in part by the National Science Fund for Distinguished Young Scholars under Grant 61925112, in part by the National Natural Science Foundation of China under Grant 61806193 and Grant 61772510, in part by the Innovation Capability Support Program of Shaanxi under Grant 2020KJXX-091 and Grant 2020TD-015 \emph{(Corresponding author: Xiangtao Zheng).}}
\thanks{Y. Zhang is with the Key Laboratory of Spectral Imaging Technology CAS, Xi'an Institute of Optics and Precision Mechanics, Chinese Academy of Sciences, Xi'an 710119, Shaanxi, P. R. China, and also with the University of Chinese Academy of Sciences, Beijing 100049, P. R. China.}
\thanks{X. Zheng and X. Lu are with the Key Laboratory of Spectral Imaging Technology CAS, Xi'an Institute of Optics and Precision Mechanics, Chinese Academy of Sciences, Xi'an 710119, Shaanxi, P. R. China. (e-mail:
xiangtaoz@gmail.com).}
}

\maketitle
%\begin{spacing}{1.25}
\begin{abstract}

Remote sensing scene classification aims to assign a specific semantic label to a remote sensing image.
Recently, convolutional neural networks have greatly improved the performance of remote sensing scene classification.
However, some confused images may be easily recognized as the incorrect category, which generally degrade the performance. The differences between image pairs can be used to distinguish image categories. This paper proposed a pairwise comparison network, which contains two main steps: pairwise selection and pairwise representation.
The proposed network first selects similar image pairs, and then represents the image pairs with pairwise representations. The self-representation is introduced to highlight the informative parts of each image itself, while the mutual-representation is proposed to capture the subtle differences between image pairs. Comprehensive experimental results on two challenging datasets (AID, NWPU-RESISC45) demonstrate the effectiveness of the proposed network. The codes are provided in https://github.com/spectralpublic/PCNet.git.
\vspace{+0.25cm}

\end{abstract}
%\end{spacing}

\begin{IEEEkeywords}
Convolutional neural networks, remote sensing scene classification, multi-branch methods.
\vspace{+0.25cm}
\end{IEEEkeywords}

\IEEEpeerreviewmaketitle

\section{Introduction}

\IEEEPARstart{R}{emote-sensing} sensing scene classification tries to classify a remote sensing image according to the scene category, which is a fundamental and important task in remote sensing image interpretation.
The remote sensing images can be described by a holistic representation \cite{zheng2020deep}, which has wide applications such as image retrieval \cite{xu2020mental}, change detection \cite{zheng2021unsupervised}, visual question answering \cite{zheng2021mutual}, object detection \cite{yao2021automatic}, and so on.
However, due to the complexity and diversity of remote sensing scenes, some confused images may be easily recognized as the incorrect category.

\par
With the rise of deep learning, Convolutional Neural Networks (CNN) can extract high-level representative features \cite{li2021error-tolerant}, which  have achieved great success in remote sensing scene classification \cite{li2021dla-matchNet}.
According to the network architecture, the CNN methods can be divided into single-branch methods as well as multi-branch methods. Single-branch methods exploit single CNN to represent the remote sensing image with convolutional feature \cite{he2020skip, wang2021multilevel} or Fully Connected (FC) feature \cite{ wei2020marginal}.
Cao \emph{et al.} \cite{cao2021self} proposed a self-attention-based deep feature fusion method for deep feature aggregation and emphasizing the weights of complex objects for remote sensing scene classification.
Sun \emph{et al.} \cite{sun2020remote} proposed a gated bidirectional network that integrates hierarchical feature aggregation and interference information cancellation into an end-to-end network.
However, single-branch methods only handle a single image, which can not consider the comparison between images.

\par
To improve representational power, multi-branch methods employ multi-branch architecture to consider some different inputs such as multi-scale of an image \cite{chen2022remote, xu2021remote}, or different images \cite{zheng2022generalized, liu2019siamese}.
Wang \emph{et al.} \cite{wang2020looking} proposed a multi-scale representation by a global local dual-branch architecture. Liu \emph{et al.} \cite{liu2019siamese} used image pairs as inputs and enhanced the robustness of the network by learning the regularization term through metrics.
Existing multi-branch methods usually compare different scales of a single image, or compare sample distributions between different image features \cite{li2021robust}. However, the difference between two images is rarely directly exploited to improve the class discriminative ability.

\begin{figure}
  \centering
  \includegraphics[width=0.88\linewidth]{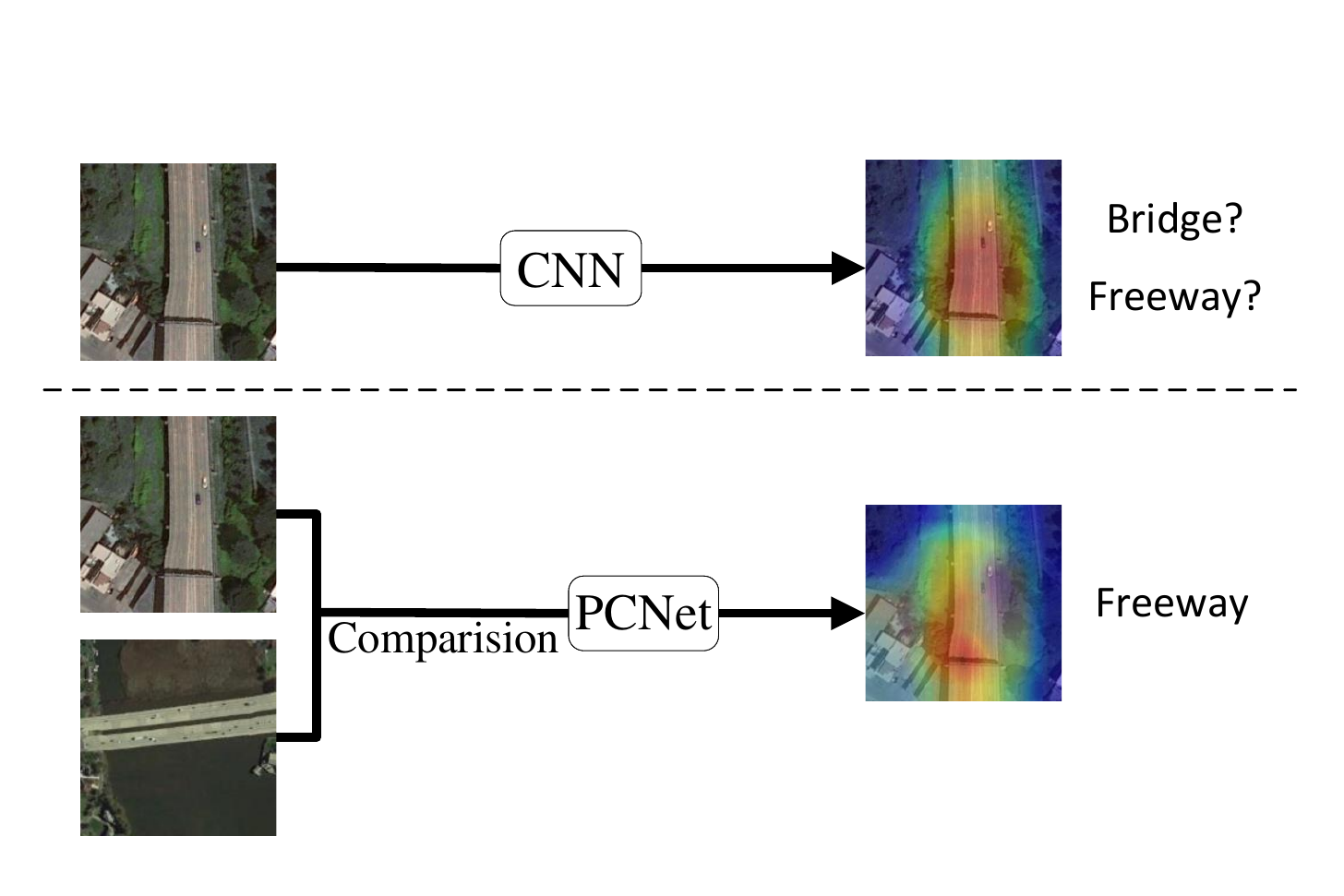}
  \vspace{0.3cm}
  \caption{An example of the difference between the normal CNN and the proposed PCNet.}\label{fig1}
  \vspace{0.3cm}
\end{figure}

\par
Given a pair of confused images, the category can be identified by comparing the subtle differences between the confused images \cite{zhuang2020learning}.
For example, the \textbf{freeway} and the \textbf{bridge} in Figure \ref{fig1}. Although the CNN has been able to gradually focus on the subject of the \textbf{freeway} through training, it is usually difficult to discriminate them because it has similar features to the \textbf{bridge}. Comparing them shows that there are differences in their backgrounds, for instance, more semantic information will be present at the two ends of the \textbf{bridge}, while the \textbf{freeway} appears on both sides.
By comparing image pairs, the difference information between images can be used to promote better classification.
The above facts show that it is easier to focus on discriminative regions by comparing image pairs, especially for confused images \cite{zhang2022pairwise}.

\par
Based on the above ideas, a pairwise comparison network (PCNet) is proposed to represent the discriminative regions by comparing image pairs.
The proposed method contains two main steps: pairwise selection and pairwise representation.
\begin{enumerate}
\item Pairwise Selection: To handle the confused image, pairwise selection is exploited to select the most similar image according to the input image. The ResNet50 is used to extract the deep features and Euclidean distance is used to measure the relationship between image pairs.
\item Pairwise Representation: To represent image pairs, two kinds of representation are considered: the self-representation and the mutual-representation. The self-representation is obtained by directly feeding the extracted features into the attention module, which focuses on the informative parts of each image itself. The mutual-representation is obtained by highlighting the comparison cues between the image pairs on each image, which captures the difference information from the perspective of individual images. Both self-representation and mutual-representation are later passed through a linear classifier for classification, while a ranking loss is introduced to consider the feature priorities: the self-representation should be more discriminative than the the mutual-representation.
\end{enumerate}

\par
Main contributions of this paper are as follows:
\begin{enumerate}
  \item To deal with the confused images in remote sensing scenes, a pairwise comparison network is proposed to capture the discriminative regions by comparing image pairs. The proposed PCNet first selects similar image pairs, and then represents the image pairs with pairwise representations.
  \item Both self-representation and the mutual-representation are considered to represent the comparing image pairs. The self-representation highlights the informative parts in each image itself and the mutual-representation captures the subtle differences between similar image pairs.
  \item A ranking loss is introduced to boost the discriminative power of both features: the self-representation should be more important than the the mutual-representation by a predefined margin. A large number of experiments have shown that the proposed method is superior to the state-of-the-art methods.
\end{enumerate}

%\vspace{-0.51cm}

\begin{figure*}[t]
  \centering
  \includegraphics[width=0.9\linewidth]{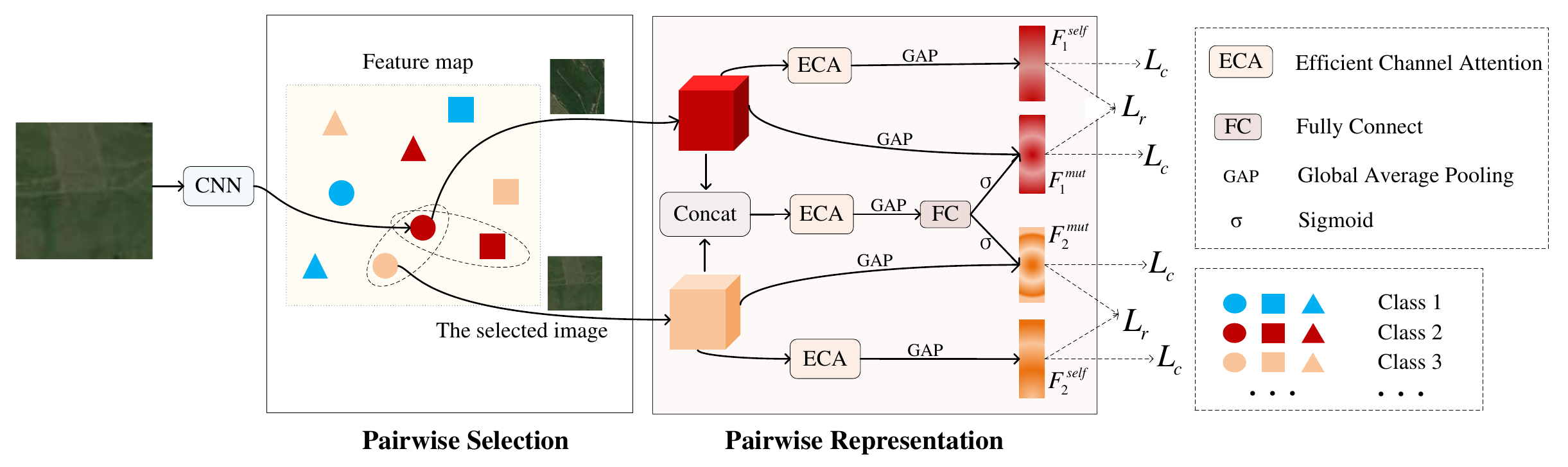}
  \vspace{0.3cm}
  \caption{The architecture of the proposed PCNet. The proposed PCNet is mainly composed by pairwise selection and pairwise representation.
The whole network is trained under the  cross-entropy loss $L_c$ and the ranking loss $L_r$.}\label{fig2}
\vspace{0.4cm}
\end{figure*}

\vspace{0.2cm}
\section{Proposed Method}\label{sec3}
\vspace{0.2cm}
\par
To distinguish the confused images, a PCNet is proposed by comparing subtle differences between image pairs. As shown in Figure \ref{fig2}, the proposed PCNet first selects similar image pairs, and then represents the comparing image pairs with pairwise representations.
\begin{enumerate}
\item \textbf{Pairwise Selection:} To form image pairs, the most similar intra-class and inter-class images are selected, and the Euclidean distance is used to measure image similarity between deep features.
\item \textbf{Pairwise Representation:} The self-representation is obtained by directly passing the obtained features through an attention module to highlight the informative parts. To obtain the mutual-representation, the obtained features of image pairs are connected to capture the comparison cues. The obtained comparison cues are then compared with the two images respectively to generate the mutual-representation from the perspective of each individual image.
The whole network is trained in an end-to-end manner with the cross-entropy loss and the ranking loss.
\end{enumerate}
%\vspace{-0.4cm}

\subsection{Single-Branch Network} \label{sec3-0}
\par
The CNN methods input an image to extract the image features, and a liner classifier for classification output. The last convolutional layer features are extracted,
%\vspace{-0.2cm}
\begin{equation}\label{1}
  \textbf{f}=f\left(\textbf{I} ; \pmb{\omega}\right),
  %\vspace{-0.2cm}
\end{equation}
where $\textbf{f}\in\mathbb{R}^{W \times H \times C}$ is the extracted features, \textbf{I} is the input image, $\pmb{\omega}$ is the parameter of pre-training CNN network. A Global Average Pooling (GAP) layer is used to connect the last convolutional layer and the FC layer,
%\vspace{-0.2cm}
\begin{equation}\label{2}
  \textbf{F}_{gap}=\frac{1}{H W} \sum_{i=0}^{H} \sum_{j=0}^{W} \textbf{f}(c, i, j).
  \vspace{0.2cm}
\end{equation}
The extracted features $\textbf{F}_{gap}\in\mathbb{R}^{C}$ are fed into a FC layer and add a softmax operation to predict the scores of each category,
%\vspace{-0.2cm}
\begin{equation}\label{3}
 \textbf{q}=\operatorname{softmax}\left(\textbf{W} \textbf{F}_{gap}+\textbf{b}\right),
 \vspace{0.2cm}
\end{equation}
where $ \textbf{q} \in\mathbb{R}^N$ is the predicted score vector, $\ N$ is the number of categories of the data, $\left\{{\textbf{W}}, {\textbf{b}}\right\}$ are the weight and bias of the FC. The cross-entropy loss function is used to train the network,
%\vspace{-0.2cm}
\begin{equation}\label{4}
  \mathcal{L}_{c}=-\sum \textbf{y}^\text{T} \log (\textbf{q}),
\end{equation}
where $\textbf{y}$ indicates the one-hot vector of the ground truth label.
By minimizes the cross-entropy loss, the distance between the ground truth and predicted probability distributions is closer, so that the trained network can obtain discriminative classification capability.
%\vspace{-0.4cm}

\subsection{Pairwise Comparison Network} \label{sec3-1}
%\vspace{-0.1cm}
\par
The single-branch network ignores the comparison between images, while the subtle differences between images are the key to distinguish confused images.
This paper constructs a pairwise comparison multi-branch network to distinguish confused images by two steps, \emph{i.e.}, pairwise selection and pairwise representation.
%As shown in Figure \ref{fig2}, the proposed PCNet first selects similar image pairs, and then represents the comparing image pairs with pairwise representations.

%by two steps, \emph{i.e.}, pairwise selection and pairwise representation.

\subsubsection{Pairwise Selection}
\par
For confused images, it is usually difficult to discern their contents by looking at one image, but humans can focus on the subtle differences between image pairs of different classes if another similar image is introduced as a comparison.
Therefore, similar images are selected as input pairs for the network.
For pairwise selection, the proposed PCNet selects the most similar inter-class and intra-class images for each image, and the image similarity is calculated based on the Euclidean distance between the extracted image features.
The comparison between intra-class image pairs can enhance the discriminative information of this category, and the comparison between inter-class image pairs can use the difference information to distinguish confused images.

\subsubsection{Pairwise Representation}
\par
The obtained features $\textbf{f}_{1} \in \mathbb{R}^{W \times H \times C}$ and $\textbf{f}_{2} \in \mathbb{R}^{W \times H \times C}$ of a image pairs feature (only one pair of inter-class similar image pairs is used as an example) can represent the content of remote sensing images in terms of global aspects.
However, they are limited to explore the target information from the images.
To further enhance the representations, this paper introduces two different representations: the self-representation which highlights the informative parts on each image itself and the mutual-representation that captures the subtle differences between image pairs.

\begin{figure}[!t]
  \centering
  \includegraphics[width=0.85\linewidth]{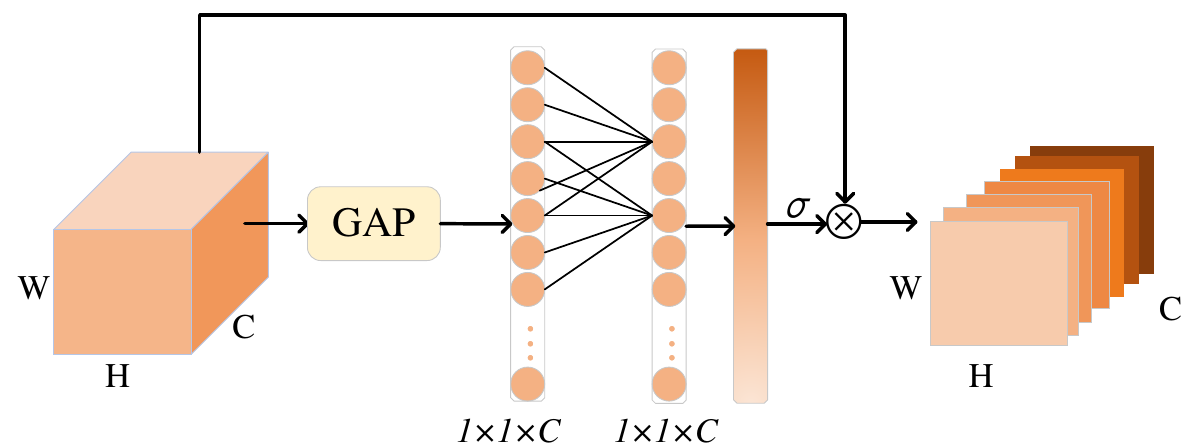}
  \vspace{0.3cm}
  \caption{The architecture of ECA model.}\label{fig3}
  %\vspace{0.2cm}
\end{figure}

\par
\textbf{Self-representation}. The extracted features $\textbf{f}_{1}$ and $\textbf{f}_{2}$ are later fed into the attention module to obtain the self-representation $\textbf{F}_{1}^{self} \in \mathbb{R}^{C}$ and $\textbf{F}_{2}^{self} \in \mathbb{R}^{C}$, thereby improving the classification performance by focusing on the discriminative regions of the image.
The attention module uses the Efficient Channel Attention (ECA) model proposed \cite{wang2020eca}, which is an improvement of the SE model \cite{hu2018squeeze}. The advantage of the ECA model over other attention models is that it introduces few additional parameters and the computational effort is largely negligible. Its architecture is shown in Figure \ref{fig3}, the extracted image features $\textbf{f}$ through GAP layer to obtain a $1\times1\times{C}$ vector, and the channel weights through performing a fast $1D$ convolution of size $k$ to capture cross-channel interactions, then the channel weights is weighted the feature to highlight the focus region.
According to the size strategy of $k$ in the original paper, $k=5$ is chosen in this paper.

\par
\textbf{Mutual-representation}. Mutual-representation focus on the existence of subtle differences between two similar images as an aid to judgment, and the main process of obtaining mutual-representation is: the extracted features $\textbf{f}_1$ and $\textbf{f}_2$ are connected and sent to the attention module to get the connected features $\textbf{f}_{cat} \in \mathbb{R}^{W \times H \times 2C}$.
In order to align $\textbf{f}_{cat}$ and $\textbf{f}_{1}$, $\textbf{f}_{cat}$ is downscaled through a FC layer.
After that, a sigmoid function is added to get $\textbf{a}_{mut}\in \mathbb{R}^{C}$, $\textbf{a}_{mut}$ represents the pairwise relations on each channel.
By comparing $\textbf{a}_{mut}$ with $\textbf{f}_1$ and $\textbf{f}_2$ separately, the different cues between the image pairs can be generated from the perspective of each individual image, which contribute to distinguishing the image pair. The obtained mutual-representations can be formulated as follows:
\vspace{0.15cm}
\begin{equation}\label{5}
  \textbf{F}_{1}^{mut}=\textbf{F}_{1} \odot \textbf{a}_{mut},
  \textbf{F}_{2}^{mut}=\textbf{F}_{2} \odot \textbf{a}_{mut},
\vspace{0.1cm}
\end{equation}
where $\odot$ denotes the product of two vectors, $\textbf{F}_{1} \in \mathbb{R}^C$ and $\textbf{F}_{2} \in \mathbb{R}^C$ are obtained by $\textbf{f}_1$ and $\textbf{f}_2$ through GAP.
It is worth noting that, unlike self-representation, $\textbf{F}_1^{mut} \in \mathbb{R}^C$ and $\textbf{F}_2^{mut} \in \mathbb{R}^C$ generate their respective representation, called mutual-representation, from the perspectives of the two images by contrast cues.
Mutual-representation prominently contains the channels of contrast cues, which facilitate the network to help classification by focusing on the local differences of the confused images.
%\vspace{-0.4cm}

\subsection{Overall Object Function}\label{sec3-2}
\par
Finally, the four obtained representation vectors $\textbf{F}_n^k\in\mathbb{R}^N$ are added to the softmax classifier to obtain the predicted score vectors $\textbf{q}_n^k\in\mathbb{R}^N$ for each category, where $n\in\{1, 2\}$, $k\in\{self, mut\}$,
%\vspace{-0.3cm}
\begin{equation}\label{6}
 \textbf{q}_{n}^{k}=\operatorname{softmax}\left(\textbf{W} \textbf{F}_{n}^{k}+\textbf{b}\right).
 \vspace{0.2cm}
\end{equation}
The classification loss function is formulated as the cross-entropy loss to minimize the distances between the predicted probabilities and the ground truth labels,
\vspace{0.2cm}
\begin{equation}\label{7}
  \mathcal{L}_{c}=-\sum_{n \in\{1,2\}} \sum_{k \in\{{self,mut}\}} \textbf{y}_{n}^{\text{T}} \log_{}{\textbf{q}_{n}^{k}}.
 \vspace{0.2cm}
\end{equation}

\par
In order to assist the judgment, the attention region of the contrast image is introduced into the image. However, since the confused image exists in the remote sensing image is only a small part, the main attention is paid to the focus region of the image itself in the judgment, and the region containing the contrast clues is only as an aid. Therefore, this paper introduces ranking loss to rank the two features generated by one image.
\vspace{0.15cm}
\begin{equation}\label{8}
  \mathcal{L}_{r}=\sum_{n \in\{1,2\}} \max \left(0, \textbf{q}_{n}^{{mut}}\left(c_{n}\right)-\textbf{q}_{n}^{self}\left(c_{n}\right)+\epsilon\right),
  \vspace{0.2cm}
\end{equation}
where $c_n$ is the corresponding index associated with the ground truth label of image $n$.
The $\mathcal{L}_{r}$ makes the score difference  $\textbf{q}_{n}^{{self}}\left(c_{n}\right)-\textbf{q}_{n}^{mut}\left(c_{n}\right)$ should be greater than a margin $\epsilon$. This allows the network to learn each image in the recognition pair adaptively considering the feature priority. Therefore, the final loss function for training the whole network is,
\vspace{0.2cm}
\begin{equation}\label{9}
  \mathcal{L}=\mathcal{L}_{c}+\lambda \mathcal{L}_{r},
  \vspace{0.2cm}
\end{equation}
where the $\lambda$ is a hyper-parameter.

\par
It is worth noting that the proposed PCNet is only used in the training phase. During training, PCNet can learn the different cues between the similar image pairs directly and thus gradually improve the discriminative ability of CNN for confused images. Therefore, during testing, it is simply offloaded and tested with the trained backbone for a single image. Specifically, a test image is directly input to the CNN backbone, and the prediction score vector is obtained by equation Eq.~(\ref{1})(\ref{2})(\ref{3}) for label prediction. PCNet is tested in the same way as a normal CNN, which largely improves the value of PCNet for remote sensing scene classification.
\vspace{0.3cm}

\section{Experiments}\label{sec4}
\vspace{0.2cm}
\subsection{Dataset and Implementation Details}
\vspace{0.2cm}
\subsubsection{Dataset}
\par
The proposed PCNet is mainly experimented on two datasets, both the AID dataset \cite{2017aid} and the NWPU-RESISC45 Dataset \cite{Cheng2017nwpu}.

AID dataset: consists of 30 scene categories with 200 to 400 images per category, the size of each image is $600 \times 600$, and has an image resolution of 0.5m to 8m per pixel.

NWPU-RESISC45 Dataset: contains 31,500 remote sensing images and 45 scene categories, each scene category contains 700 images with a size of $256 \times 256$, the images have a spatial resolution of 30m to 0.2m per pixel.

The ratio of training sets is set to 20\% and 50\% for AID dataset and 10\% and 20\% for NWPU-RESISC45 Dataset. To make full use of the training data and avoid overfitting, a data enhancement strategy is used by randomly rotating the training images by $30^{\circ}$, flipping them horizontally, and flipping them vertically. Overall accuracy (OA) is used evaluation metrics, where OA refers to the ratio of the number of correct images classified during testing to the total number of images tested.

\subsubsection{Training Detail}
\par
This experiment is run on a computer with GTX TITAN X 12G GPU and implemented by the Pytorch. The backbone of CNN selected in this paper is ResNet50 which pre-trained on ImageNet. The specific parameter configuration is shown below: The epoch of the model is 100, using a standard SGD with a momentum of 0.9 and a weight decay of 0.0005. The initial learning rate is 0.01 and is adjusted using an adopt cosine annealing strategy. When constructing the image pairs, the most similar images between intra-class and inter-class in each batch are selected by Euclidean distance of the extracted image features.

\par
For each batch, 30 classes of images are selected in the dataset and 6 images are randomly selected for each class. This paper explored how to set the number of images most reasonably on AID dataset for 20\% traning set. As shown in Table \ref{table1}, fixing the category and the number of images per category better than random selection, it can be inferred that fixing the category and the number of images is more beneficial for pairwise selection.
Among them, the proposed PCNet works best when 6 images of each categories are selected for each batch.

\par
As shown in Table \ref{table2}, compared to the ResNet50, the proposed PCNet obtains + 2.3\% OA, adds about 9M parameters number and 0.7G Floating Point Operations Per Second (FLOPs) in training, which increases the cost of mainly four representations. More additional computations are required for the network to focus on the image focus areas of interest as well as the areas where similar images have differences.

%\vspace{-0.4cm}

\subsection{Ablation Experiments}
\vspace{0.2cm}
\par
To investigate the effectiveness of the proposed method, this paper evaluated its key designs on the AID dataset with a data division of 20\% for training. To be fair, when evaluating a design with different strategies, the other settings are kept consistent with those in the implementation details.

\begin{table}[!t]
  \renewcommand{\arraystretch}{1.1}
% if using array.sty, it might be a good idea to tweak the value of
% \extrarowheight as needed to properly center the text within the cells
  \caption{Results of Batch Selection Method}
  \label{table1}
  \centering
  \begin{tabular}{c c|c }
    \hline
    %Bdtch & ~ & OA\\
    \multicolumn{2}{c|}{Batch} & \multirow{2}*{OA\%}\\
    \cline{1-2}
    Classes & Image & ~\\
    \hline
    %\multicolumn{2}{c|}{{\color{blue}Randomly selected 180 images}} & 94.61\\
    \multicolumn{2}{c|}{Randomly selected 180 images} & 94.61\\
    30 & 4 & 95.15\\
    30 & 6 & \textbf{95.53}\\
    30 & 8 & 95.25\\
    \hline
    \vspace{0.2cm}
  \end{tabular}

\end{table}

\begin{table}[!t]
%\vspace{-0.2cm}
 \renewcommand{\arraystretch}{1.1}
% if using array.sty, it might be a good idea to tweak the value of
% \extrarowheight as needed to properly center the text within the cells
  \caption{Computation Cost Analysis}
  \label{table2}
  \centering
  \begin{tabular}{c|c|c|c}
    \hline
    Method & Parameters & FLOPs & OA\%\\
    \hline
    ResNet50 & 23.6M & 3.86G &93.24\\
    PCNet& 32.1M & 3.87G &\textbf{95.53}\\
    \hline
  \end{tabular}
  \vspace{0.3cm}
\end{table}

\begin{table}[!t]

  \renewcommand{\arraystretch}{1.1}
% if using array.sty, it might be a good idea to tweak the value of
% \extrarowheight as needed to properly center the text within the cells
  \caption{Comparison of Different Components}
  \label{table3}
  \centering
  \begin{tabular}{c|c|c|c|c}
    \hline
    Methods&Architecture&Representation&Object function&OA\%\\
    \hline
       Baseline&Single-branch &  Self & $L_c$ & 94.71\\
    %$\text{Multi}_1$ &\multirow{4}{*}{Multi-Branch} &  self& $L_c$ & 94.98\\
   $\text{Multi}_1$ &Multi-branch &  Self& $L_c$ & 94.98\\
   $\text{Multi}_2$ &Multi-branch &  Mutual & $L_c$ & 94.82\\
   $\text{Multi}_3$ &Multi-branch &  Self+mutual & $L_c$ & 95.09\\
    PCNet &Multi-branch & Self+mutual& $L_c$+$L_r$ & \textbf{95.53}\\
    \hline
  \end{tabular}
\vspace{0.3cm}
\end{table}

\begin{table}[!t]
  \renewcommand{\arraystretch}{1.1}
% if using array.sty, it might be a good idea to tweak the value of
% \extrarowheight as needed to properly center the text within the cells
  \caption{Results of Different selection methods}
  \label{table4}
  \centering
  \begin{tabular}{c|c}
    \hline
    Selection methods  & OA\%\\
    \hline
    Random & 95.18\\
    Cosine similarity& 95.25\\
    European distance & \textbf{95.53}\\
    \hline
  \end{tabular}
  \vspace{0.2cm}
\end{table}

\begin{table}[!t]
%\vspace{-0.2cm}
  \renewcommand{\arraystretch}{1.1}
% if using array.sty, it might be a good idea to tweak the value of
% \extrarowheight as needed to properly center the text within the cells
  \caption{Results of Construction of Image Pairs}
  \label{table5}
  \centering
  \begin{tabular}{c c |c}
    \hline
    Inter & Intra  & OA\%\\
    \hline
    Random &Random & 95.18\\
    S & Random & 95.32\\
    S & D & 95.21\\
    S & S & \textbf{95.53}\\
    \hline
  \end{tabular}
  \vspace{0.3cm}
\end{table}

\begin{table}[!t]
 \renewcommand{\arraystretch}{1.1}
% if using array.sty, it might be a good idea to tweak the value of
% \extrarowheight as needed to properly center the text within the cells
  \caption{Analysis of Hyper-Parameter $\lambda$}
  \label{table6}
  \centering
  \begin{tabular}{c|ccccc}
    \hline
    $\lambda$ & 0.5 & 0.8 & 1.0 & 1.2 & 1.5\\
    \hline
    OA\% & 95.26 & 95.37 &\textbf{95.53} & 95.33 &95.11\\
    \hline
  \end{tabular}
  \vspace{0.3cm}
\end{table}

\begin{figure}[!t]
  \centering

  \includegraphics[width=0.8\linewidth]{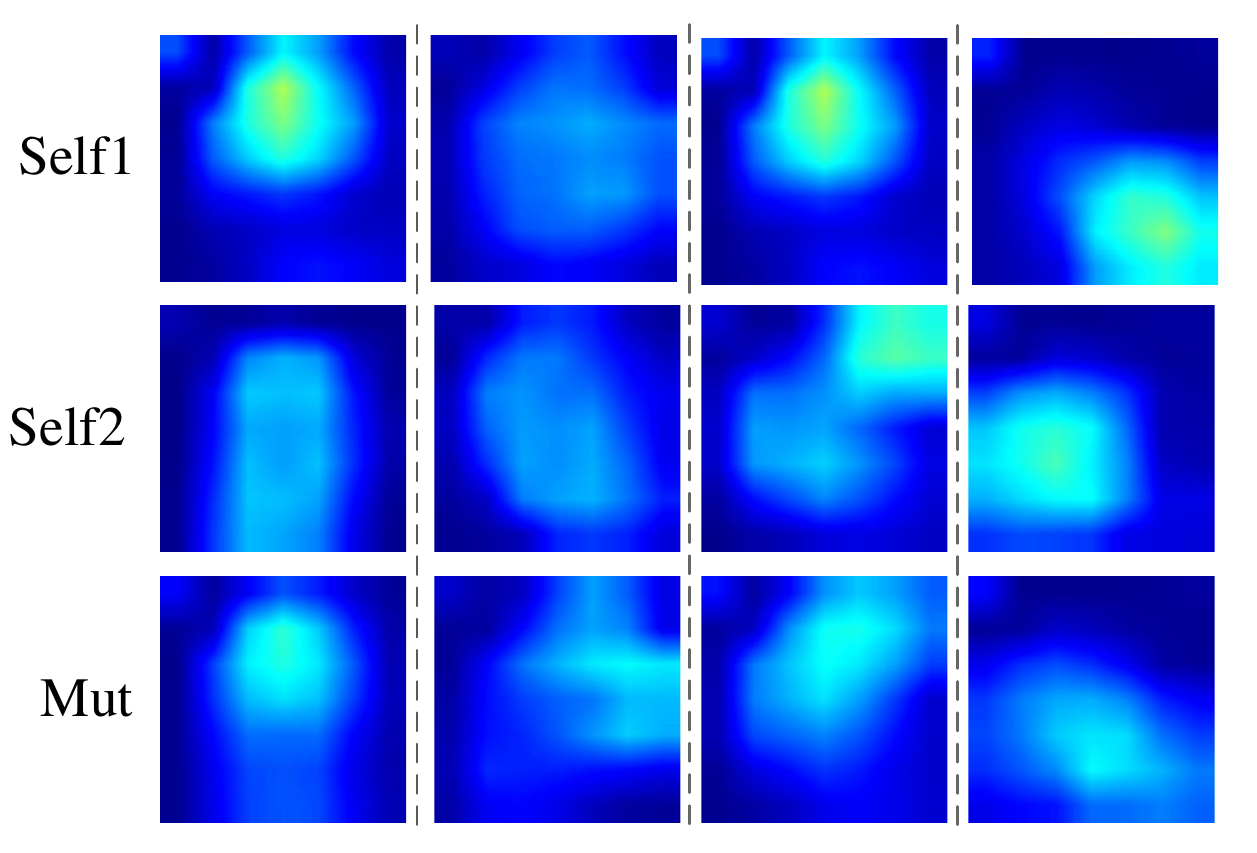}
  \caption{The Visualization of the Difference Features.}\label{fig4}
  %\vspace{-0.3cm}
\end{figure}

\begin{table*}[!t]
  \renewcommand{\arraystretch}{1.1}
% if using array.sty, it might be a good idea to tweak the value of
% \extrarowheight as needed to properly center the text within the cells
  \caption{Comparison with the State-of-the-Art Methods}
  \label{table7}
  \centering
  \begin{tabular}{c|c|cc|cc}
   \hline
   \multirow{2}{*}{Architecture} &\multirow{2}{*}{Methods} & \multicolumn{2}{c|}{AID dataset (OA\%)} & \multicolumn{2}{c}{NWPU-RESISC45 dataset (OA\%)} \\
                         &  & 20\% for training       & 50\% for training       & 10\% for training        & 20\% for training       \\
   \hline
    \multirow{5}{*}{Single-branch} & SCCov \cite{he2020skip}& 93.12$\pm$0.25  & 96.10$\pm$0.16  & 89.30$\pm$0.35  & 92.10$\pm$0.25      \\
    & ACR-MLFF \cite{wang2021multilevel}& 92.73$\pm$0.12   & 95.06$\pm$0.33  & 90.01$\pm$0.33  & 92.45$\pm$0.20     \\
    &IORN+marginal center loss \cite{wei2020marginal}& 94.05$\pm$0.12   & - & 89.97$\pm$0.15  & 92.95$\pm$0.09     \\
    &SAFF \cite{cao2021self}& 90.25$\pm$0.29  & 93.83$\pm$0.28   & 84.38$\pm$0.19   & 87.86$\pm$0.14       \\
    & GBNet \cite{sun2020remote}& 92.20$\pm$0.23     & 95.48$\pm$0.12   &     -   &         -      \\
    & ResNet50+EAM \cite{zhao2020remote}& 93.64$\pm$0.25  & 96.62$\pm$0.13    & 90.87$\pm$0.15   & 93.51$\pm$0.12     \\
   \hline
   \multirow{3}{*}{Multi-branch}
    & SKAL \cite{wang2020looking}& 94.38$\pm$0.10   & 96.76$\pm$0.20    & 90.04$\pm$0.15   & 92.79$\pm$0.11        \\
    & GLDBS \cite{xu2021remote}& 95.45$\pm$0.19 & \textbf{97.01$\pm$0.22} & 92.24$\pm$0.21    & 94.46$\pm$0.15   \\
    & PCNet               & \textbf{95.53$\pm$0.16}         & 96.76$\pm$0.25       & \textbf{92.64$\pm$0.13}          & \textbf{94.59$\pm$0.07}\\
   \hline
 \end{tabular}
\vspace{0.2cm}
\end{table*}

\subsubsection{Different Components}
To verify the effectiveness of the proposed PCNet, three key components are evaluate: architecture, representation and object function.
\begin{enumerate}
\par
\item To show the effect of multi-branch networks, Table \ref{table3} presents two architectures with the same ResNet50 and attention module: baseline method and $\text{Multi}_1$ method. The baseline method exploits a single branch network to extract self-representation from a single image, while $\text{Multi}_1$ method exploits multi-branch networks to extract self-representation from two images.
\item To demonstrate the necessity of the pairwise representation, three different representations networks are explored with the same architecture and object function: $\text{Multi}_1$, $\text{Multi}_2$, and $\text{Multi}_3$. $\text{Multi}_1$ method uses self-representation, $\text{Multi}_2$ method uses mutual-representation, and $\text{Multi}_3$ method uses both self-representation and mutual-representation.
\item To show the effect of the ranking loss, $\text{Multi}_3$ is trained only with cross entropy loss function, the proposed PCNet is trained with the object function $L_c$+$L_r$. As shown in Table \ref{table3}, the proposed PCNet achieves the best performance.
\end{enumerate}

\subsubsection{Pairwise Selection}
To verify the effectiveness of the pairwise selection, this paper evaluates the different selection methods and construction of image pairs methods.

\begin{enumerate}
\item To pick the most similar image pairs, this paper explores the effect of random selection, cosine similarity, and Euclidean distance on the performance of PCNet. Table \ref{table4} shows Euclidean distance is effective to pick similar image pairs.

\item The construction strategies for different image pairs are explored in Table \ref{table5}, where \textbf{D} denotes the most different image pair and \textbf{S} denotes the most similar image pair. It shows that the best results are obtained when both intra-class and inter-class images are the most similar. This indicates that the proposed PCNet is more likely to discriminate from similar images by focusing on the differences between image pairs. Training the most similar image pairs between intra-class and inter-class, the proposed PCNet can gradually distinguish which are really confused images or truly similar image pairs.
\end{enumerate}
\subsubsection{ Analysis of Hyper-Parameter $\lambda$}
\par
As shown in Eq.~(\ref{9}), the proposed PCNet involves a hyper-parameter $\lambda$, where the effect of $\lambda$ on the whole network is evaluated in Table \ref{table6}. The best results were obtained when $\lambda=1$, confirming the effectiveness and importance of introducing ranking loss.
%\vspace{-0.4cm}

\subsection{Comparison with the State-of-the-Art Methods}
\vspace{0.2cm}
\par
In order to verify the effectiveness of the proposed method, the state-of-the-art methods are shown in Table \ref{table7}. From the results, it can be concluded that the multi-branch methods is more effective than the single-branch methods. And among the multi-branch methods, the effectiveness of the proposed PCNet is better.
For the AID dataset, when the training data is 50\%, the PCNet achieves 96.76\%$\pm$0.25 accuracy which is only about 0.24\% lower than the current best-performing method (GLDBS \cite{xu2021remote}). This is because that GLDBS combines the two backbones to provide a significant advantage, where the proposed PCNet is a single backbones. Despite that, the PCNet always achieves better performance than GLDBS when the training data is 20\%. This comprehensive comparison demonstrates that the PCNet can helps the network to discriminate the confused images in remote sensing images, thus enhance the performance of the deep network for remote sensing scene classification.

\par
Figure \ref{fig4} shows the visualization of the model response of different features. The \textbf{self1} and \textbf{self2} represent the visualization obtained by the the features $\textbf{f}_1$ and $\textbf{f}_2$ through the attention module, and the \textbf{Mut} represents the visualization of the $\textbf{f}_{cat}$. It can be seen the focus areas of image pairs are different, and pairwise comparison will guide the network to focus on more discriminative areas of the image pairs to help judgment.
%GLDBS combines ResNet18 and ResNet34 to learn global and local information. Even though the PCNet exploits a single backbone, it always achieves better performance than GLDBS when the training data is 20\%.

\vspace{0.3cm}

\section{Conclusion}\label{sec5}
\vspace{0.25cm}
\par
This paper has proposed a PCNet for remote sensing scene classification to solve the problem that remote sensing images have a large number of confused images which are difficult to discriminate.
The proposed PCNet firstly selects image pairs based on the Euclidean distance, then jointly exploits two different representations to distinguish the image pairs.
Specifically, the self-representation is introduced to highlight the informative regions of each image itself, and the mutual-representation is proposed to focus on the subtle differences between the image pairs.
By focusing on the focal regions of images and the subtle differences between similar images, the proposed PCNet achieves good results.
In future work, we will further explore more effective methods to solve confused images in remote sensing scene classification.
\vspace{0.2cm}

%---------------------------------------------------------------------

\ifCLASSOPTIONcaptionsoff
  \newpage
\fi
%\clearpage
\bibliographystyle{IEEEtran}
\bibliography{PCNet}

\end{document}